\documentclass[conference]{IEEEtran}
\IEEEoverridecommandlockouts

\usepackage{cite}
\usepackage{amsmath,amssymb,amsfonts}
\usepackage{algorithmic}
\usepackage{graphicx}
\usepackage{textcomp}
\usepackage{xcolor}
\usepackage{bm}
\usepackage{booktabs}
\usepackage{multirow}
\usepackage{makecell}
\usepackage{url}

\def\BibTeX{{\rm B\kern-.05em{\sc i\kern-.025em b}\kern-.08em
    T\kern-.1667em\lower.7ex\hbox{E}\kern-.125emX}}
\begin{document}

\title{PEGNet: A Physics-Embedded Graph Network for \\Long-Term Stable Multiphysics Simulation}

\author{
\IEEEauthorblockN{Can Yang,
Zhenzhong Wang,
Junyuan Liu,
Yunpeng Gong,
Min Jiang}
\IEEEauthorblockA{School of Informatics, Xiamen University, Xiamen, China \\
Email: minjiang@xmu.edu.cn (Corresponding author)}
}

\maketitle

\begin{abstract}
Accurate and efficient simulations of physical phenomena governed by partial differential equations (PDEs) are important for scientific and engineering progress. While traditional numerical solvers are powerful, they are often computationally expensive. Recently, data-driven methods have emerged as alternatives, but they frequently suffer from error accumulation and limited physical consistency, especially in multiphysics and complex geometries.
To address these challenges, we propose PEGNet, a Physics-Embedded Graph Network that incorporates PDE-guided message passing to redesign the graph neural network architecture. By embedding key PDE dynamics like convection, viscosity, and diffusion into distinct message functions, the model naturally integrates physical constraints into its forward propagation, producing more stable and physically consistent solutions.
Additionally, a hierarchical architecture is employed to capture multi-scale features, and physical regularization is integrated into the loss function to further enforce adherence to governing physics. We evaluated PEGNet on benchmarks, including custom datasets for respiratory airflow and drug delivery, showing significant improvements in long-term prediction accuracy and physical consistency over existing methods.
Our code is available at \url{https://github.com/Yanghuoshan/PEGNet}.
\end{abstract}


\section{Introduction}
\label{sec:intro}

Multiphysics simulations often use PDEs to model the spatio-temporal evolution of physical quantities, generating time series from numerical solutions at specific spatial points to reveal system dynamics.
Increasingly, data-driven methods are emerging as alternatives to traditional numerical solvers. Once trained, these methods can offer rapid inference with modest computational cost, overcoming the high resource demands and long runtimes of conventional approaches.

Notable advances such as Physics-Informed Neural Networks (PINNs) \cite{raissi2019physics}, Fourier Neural Operators (FNOs) \cite{lifourier}, Graph Neural Networks (GNNs) \cite{battaglia2018relationalinductivebiasesdeep}, Neural ODEs \cite{chen2018neural},  
Transformer-based models \cite{litransformer, hanpredicting}, and Diffusion-based solvers \cite{zhou2025text2pdelatentdiffusionmodels, li2024learning} have demonstrated impressive results in modeling physical processes, offering both accuracy and efficiency.
Among these, GNNs, particularly the Graph Network-based Simulator (GNS) framework \cite{sanchez2020learning}, have shown great promise in simulating a wide range of physical systems due to their ability to handle irregular structures.
The advantages of the GNS framework inspired us to apply it in the field of medical physics simulations, as this domain typically involves complex human geometries and multiphysics. Furthermore, this domain still requires differentiable models that enable rapid simulations and support optimization frameworks to optimize therapeutic strategies \cite{salarpour2025advancing}.

However, deep learning models for long-term prediction in time-dependent physical systems primarily face two challenges. 
First, autoregressive strategies, used by many models \cite{sanchez2020learning, lifourier}, accumulate small errors over time, which can cause predictions to drift from physical reality \cite{yang2025longtermautoregressivepredictionusing}. 
Second, they often neglect underlying physics. For instance, when numerically solving incompressible Navier-Stokes equations, velocity requires time integration, while pressure is directly solved and has no time evolution \cite{patankar1983calculation}.
Many models overlook this, treating all physical quantities uniformly and attempting to learn a time evolution for pressure, which is physically unsound. This limits physical consistency and worsens error accumulation, particularly in multiphysics systems.

\paragraph{Main Idea}
To address these issues, we introduce the PDE-guided message passing approach, which directly embeds the mathematical and physical semantics of governing equations into the GNN's architecture.
This approach imposes a strong inductive bias by incorporating physical priors into the model's forward propagation, guiding it to naturally follow physical constraints and produce more physically plausible solutions.
This ensures that every prediction step is guided by physical priors, thereby reducing the accumulation of errors caused by physical inconsistencies in long-term prediction. 
Furthermore, PEGNet designs separate forward propagation pathways for different types of physical variables, ensuring each quantity is handled according to its true physical nature.
This architecture significantly enhances both prediction accuracy and physical consistency.

\begin{figure*}[t]
    \centering
    \includegraphics[width=0.95\textwidth]{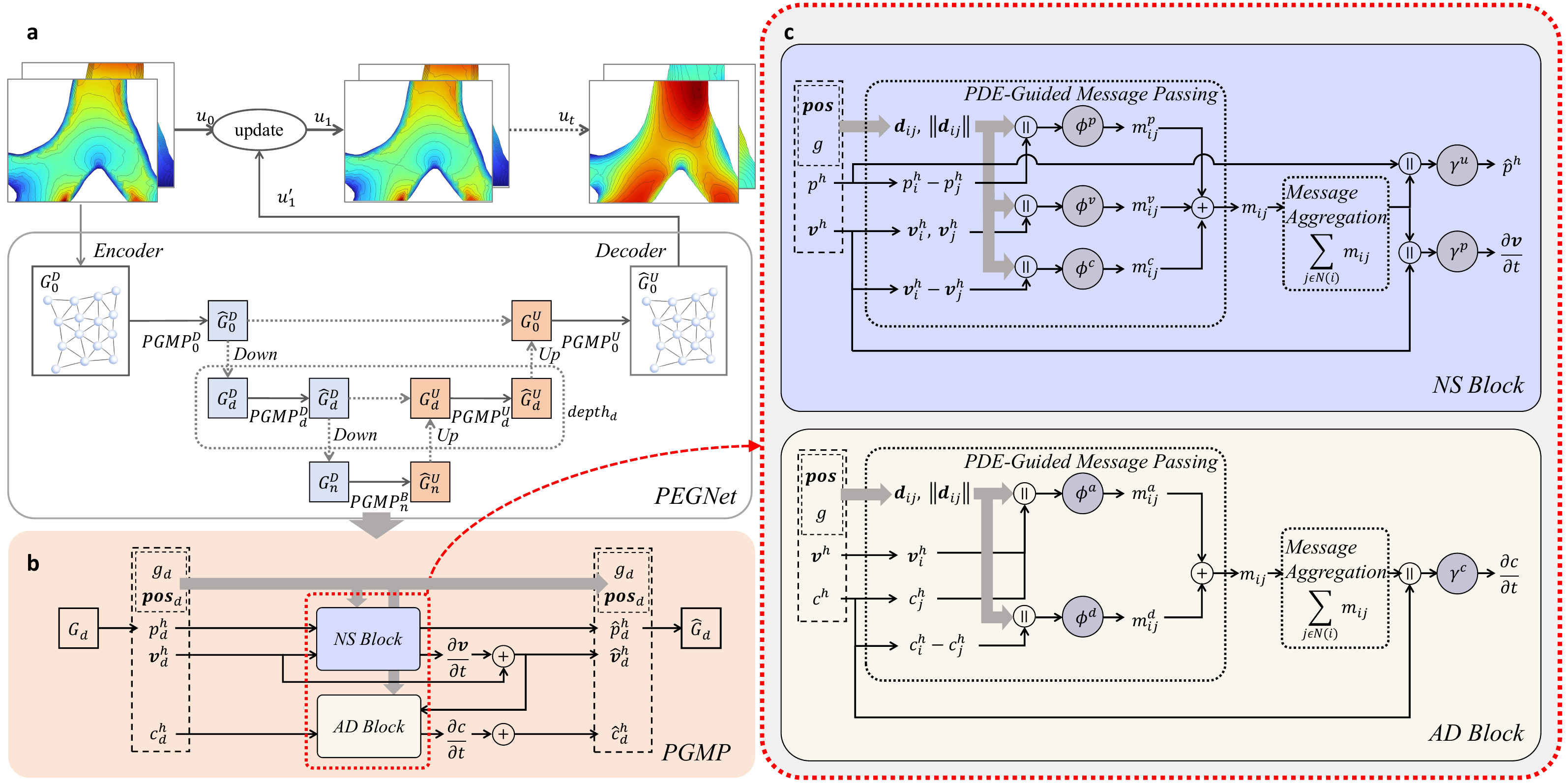}
    \caption{(a) The overall framework of PEGNet. (b) The internal structure of the PGMP module, which consists of a one-way coupled NS Block and AD Block. (c) A detailed view of the internal structures of the NS Block and AD Block, showing how PDE-guided message passing is implemented.}
    \label{fig:architecture}
\end{figure*}

\paragraph{Contributions}
We summarize the main contributions of this work as follows:
\begin{itemize}
	\item To address the issues of error accumulation and physical inconsistency, we propose PEGNet, a Physics-Embedded Graph Network with PDE-guided message passing, which naturally integrates physical constraints into the model architecture.
    \item We've built respiratory simulation datasets with two components for benchmarking single-phase flow and multiphysics drug delivery prediction: one for airflow dynamics and another for drug delivery.
    \item Our evaluation, based on long-term errors and proposed physical consistency metrics, demonstrates that PEGNet not only reduces error accumulation in long-term rollouts but also exhibits superior physical consistency compared to purely data-driven models.
\end{itemize}

\section{Related Work}
\label{sec:relatedwork}

\paragraph{GNNs for Physical Simulation}
Pioneering works from Interaction Networks \cite{battaglia2016interaction} to GNS \cite{sanchez2020learning} and MGN \cite{pfaff2020learning} have shown that GNNs provide a natural framework for learning physical dynamics on irregular domains such as particle systems or unstructured meshes.
These methods have been successfully applied in rigid-body physics \cite{yang2025mbds}, fluid dynamics \cite{gao2024predicting}, and elasticity \cite{deshpande2024magnet}.
However, most GNN-based methods adopt purely data-driven message passing, neglecting physical constraints. 
This limitation results in error accumulation and violations of physical consistency over long-term predictions.

\paragraph{Hierarchical GNN Architectures}
Hierarchical GNNs address long-range dependencies, resolution heterogeneity, and computational cost by constructing multi-scale graph representations via pooling or coarsening. Methods such as DiffPool \cite{ying2018hierarchical} and TopK pooling \cite{cangea2018sparsehierarchicalgraphclassifiers} have proven effective in graph compression.
In physical modeling, approaches like SGUNET \cite{shen2025transferlearningscalablegraph} and BSMS-GNN \cite{cao2023efficient} employ non-learnable pooling to improve generalization. These designs are especially useful for complex domains like anatomical airways, where the meshes are numerous and multi-scale.

\paragraph{Physics-Inspired Learning}
PINNs \cite{raissi2019physics} incorporate PDE residuals into loss functions, while operator learning methods such as DeepONets \cite{Lu_2021} and FNOs \cite{lifourier} aim to directly learn mappings between function spaces. Physics-informed operator networks \cite{goswami2023physics} and physics-encoded or physics-embedded approaches \cite{rao2023encoding, guo2021physics} further embed physical constraints into model design, with ongoing work exploring higher-order integration schemes \cite{wang2025multipdenetpdeembeddedlearningmultitimestepping}.
To handle irregular geometries and large-scale data, several studies combine these physics-based approaches with GNNs \cite{horie2022physics, zeng2025phympgnphysicsencodedmessagepassing}, achieving promising results in single-phase flow modeling but requiring further development for multiphysics scenarios.

\section{Proposed Method}
\label{sec:method}

\subsection{Overview of PEGNet}

PEGNet is a data-driven simulator designed to model multiphysics systems with explicit incorporation of physical priors from governing partial differential equations (PDEs). As illustrated in Figure \ref{fig:architecture}(a), the architecture follows a general Encode-Process-Decode \cite{battaglia2018relationalinductivebiasesdeep} paradigm, widely adopted in learned physical simulators \cite{sanchez2020learning, pfaff2020learning}, while incorporating physics-structured message passing mechanisms and a multi-scale structure.
Given an initial state graph with node features $u_0$ (e.g., velocity, pressure, position), PEGNet performs autoregressive rollout over $n$ time steps. At each step $t$, the model predicts the output node features $u^{\prime}_t$, and recursively feeds them back to update the input state $u_{t-1}$ for the next step $u_t$. This enables long-term simulation over the full time window $(t_0, t_0 + n\Delta t]$.
The model architecture consists of three main modules:

\paragraph{Encoder}
A multi-layer perceptron (MLP) maps the input graph with physical node features $u_t$ into a latent graph representation $G_d^D$ in the feature space. The encoded features include latent physical quantities along with positional and topological information.
For more detailed input feature information, please refer to Appendix A.

\paragraph{Processor}
Instead of using generic message passing layers \cite{gilmer2017neural}, we propose the PDE-guided message passing approach. To instantiate this approach in our model, we design a Physics-Guided Message Passing (PGMP) module that consists of two components: the Navier-Stokes (NS) Block and the Advection-Diffusion (AD) Block as shown in Figure \ref{fig:architecture}(b). 
Each block follows a specific PDE and is then coupled within the PGMP module.
Their detailed structures are presented in the following subsections.
These blocks act on the hierarchical graph $G_d$ and output updated latent graph $\hat{G}_d$, indicating that PGMP module operates on a particular resolution level $d$. Multiple modules are stacked with downsampling and upsampling layers, enabling the effective fusion of local and global physical dynamics.

\paragraph{Decoder}
A final MLP decodes the processed latent graph into target node outputs $u^{\prime}_t$, depending on the specific modeling task.
Detailed information on the output node features is included in Appendix A.

\subsection{PDE-Guided Message Passing}
\label{pde_to_gnn}

The neural network in Neural ODEs \cite{chen2018neural} can be specifically represented as follows: 
\begin{equation}
	\label{eq:NeuralODE}
	\frac{du(t)}{dt} = f_\theta(t, u(t)).
	\end{equation}
Inspired by this, we can solve the time partial derivative part in the PDE according to this formula, while using the neural network to map the remaining parts. Consider the following PDE:
\begin{equation}
	\label{eq:PDE}
	\frac{\partial u}{\partial t} = F\left(t, \bm{x}, u, \nabla u, \nabla^2 u, \dots\right),
\end{equation}
where $u(\bm{x},t) \in R^n$ is the state tensor representing $n$ physical quantities, and $\nabla^i u$ denotes the $i$-th order derivative of $u$.
The PDE-guided message passing approach establishes a structural mapping from the right-hand side of Equation \ref{eq:PDE} to the message functions of a GNN. 

We discretize the continuous PDE system onto a set of interacting nodes, forming a graph $G=(V, E)$, where each node $i \in V $ represents a fluid or material element with features $u_i$ and $\mathcal{N}(i)$ denotes the set of neighbor nodes of node $i$. 
The message $m_{ij}$ passing from nodes $j$ to $i$ is formulated to approximate the PDE terms that govern their interaction. 
Accordingly, the PDE-guided message passing approach can be specified as follows:
\begin{equation}
	\label{eq:PDEGUIDEMP}
	\begin{split}
		m_{ij} &= \sum_{r=1}^{R} \phi_r\left(z_{ij}^{(r)}\right),\\
		\frac{\partial u_i}{\partial t} = \gamma(u_i, &\text{AGG}(\{ m_{ij} : j \in \mathcal{N}(i) \})),
	\end{split}
\end{equation}
where each $z_{ij}^{(r)} \in \{ u_i, u_j, \nabla u_i, \nabla^2 u_i, \dots \}$ denotes the $r$-th component of the input features (e.g., raw values, gradients, or higher-order derivatives), and each $\phi_r$ is an independent mapping function (e.g., an MLP).
The aggregation (AGG) is performed via summation, and $\gamma$ is another mapping function responsible for calculating the time partial derivative.

Instead of learning generic, black-box interactions, we design a specific message passing scheme to explicitly compute each term of the PDE in the latent space. This approach introduces a strong inductive bias that guides the GNN to capture physical dynamics consistent with the underlying governing equations.
As illustrated in Figure \ref{fig:architecture}(b), this is realized through two specialized blocks within the PGMP Module: the NS Block and the AD Block.

\paragraph{NS Block for Fluid Dynamics}

The NS Block is designed to model fluid flow dynamics, governed by the incompressible Navier-Stokes equations:
\begin{equation}
\label{eq:NS}
	\begin{split}
		\frac{\partial \bm{v}}{\partial t} + (\bm{v} \cdot \nabla)\bm{v} &= -\frac{1}{\rho}\nabla p + \nu \nabla^2 \bm{v}, \\
		\nabla \cdot \bm{v} &= 0,
	\end{split}
\end{equation}
where $\bm{v}$ denotes the fluid velocity field, $\rho$ is the fluid density and $\nu$ is the kinematic viscosity. 
The first line of Equation \ref{eq:NS} presents the momentum equation, which, from left to right, consists of the time derivative term, the advection term, the pressure gradient term, and the viscous term. The second line imposes the divergence-free constraint.
Following Equation \ref{eq:PDEGUIDEMP}, we map each PDE term in Equation \ref{eq:NS}, excluding the partial time derivative, to a specific message function computed by a dedicated MLP.
Consequently, message $m_{ij}$ is decomposed into the pressure gradient message $m_{ij}^{p}$, the viscosity message $m_{ij}^{v} $, and the advection message $m_{ij}^{c}$.
The specific PDE-guided message passing process of the NS Block is thus formulated as follows:

\begin{equation}
    \label{eq:NSBlock}
        \begin{split}
            \quad &m_{ij}^{p} = \phi^p([p_i^h - p_j^h, \bm{d}_{ij},\|\bm{d}_{ij}\|]), \\
            \quad &m_{ij}^{v} = \phi^v([\bm{v}_i^h - \bm{v}_j^h, \|\bm{d}_{ij}\|]), \\
            \quad &m_{ij}^{c} = \phi^c([\bm{v}_i^h, \bm{v}_j^h, \bm{d}_{ij}, \|\bm{d}_{ij}\|]), \\
            \quad &\frac{\partial \bm{v}_i}{\partial t} = \gamma^u([\bm{v}_i^h, \sum_{j\in\mathcal{N}(i)}(m_{ij}^{p} + m_{ij}^{v} + m_{ij}^{c})]), \\
            \quad &\hat{p}_i^h = \gamma^p([p_i^h, \sum_{j\in\mathcal{N}(i)}(m_{ij}^{p} + m_{ij}^{v} + m_{ij}^{c})]).
        \end{split}
\end{equation}

Equation \ref{eq:NSBlock} provides a detailed illustration of the mapping between physical terms and neural components, while the structural diagram is shown in Figure \ref{fig:architecture}(c).
In this block, the pressure gradient term, which drives fluid flow from regions of high to low pressure, is approximated using pressure differences between neighboring nodes.
For incompressible flows, the fluid density $\rho$ remains constant. Thus, the coefficient term $-\frac{1}{\rho}$ can be omitted for simplicity.
The displacement vector $\bm{d}_{ij} = \bm{pos}_j - \bm{pos}_i$ captures the relative position between nodes and enables the network to learn a discretized gradient operator. The MLP learns to scale the pressure difference by this displacement, effectively approximating the gradient term $-\nabla p_i$.
Similarly, the viscous and convective terms are modeled by separate MLPs, $\phi^v$ and $\phi^c$, which generate the viscosity and convection messages, respectively, using the same principle.
Then,  the pressure, viscosity, and convection messages from neighboring nodes are aggregated by summation.
Finally, another MLP, $\gamma^u$, processes the aggregated messages along with the original velocity to predict the latent time derivative, completing the modeling of the momentum equation.

Subsequently, a separate MLP, $\gamma^p$, is used to directly predict the updated pressure field $\hat{p}_i^h$, which is crucial for enforcing the incompressibility constraint.
As discussed in Section \ref{sec:intro}, traditional numerical solvers do not evolve pressure via time integration. Instead, they determine it by solving the pressure Poisson equation, which enforces incompressibility, ensures mass conservation, and stabilizes local flow behavior. This approach contrasts with the velocity update, which typically involves explicit time integration.

\paragraph{AD Block for Scalar Transport}
The AD Block models the transport of a scalar quantity $c$, within the velocity field $\bm{v}$. This process is governed by the Advection-Diffusion equation:
\begin{equation}
\frac{\partial c}{\partial t} + \nabla \cdot (c\bm{v}) = \nabla \cdot (D \nabla c),
\label{eq:ad}
\end{equation}
where $D$ is the diffusion coefficient.
Assuming an incompressible flow ($\nabla \cdot \bm{v} = 0$), the equation simplifies to:
\begin{equation}
\frac{\partial c}{\partial t} + \bm{v} \cdot \nabla c = D \nabla^2 c.
\label{eq:ad_simple}
\end{equation}
In Equation \ref{eq:ad_simple}, from left to right, the second term is the advection term, and the third term is the diffusion term.
Similar to the NS Block, we present the structural diagram of the AD Block in Figure \ref{fig:architecture} and its detailed message passing process is as follows:
\begin{equation}
	\label{eq:ADBlock}
	\begin{split}
		\quad &m_{ij}^{a} = \phi^{a}([\bm{v}_i^h, c_j^h - c_i^h, \bm{d}_{ij}]), \\
		\quad &m_{ij}^{d} = \phi^{d}([c_j^h - c_i^h, \|\bm{d}_{ij}\|]), \\
		\quad &\frac{\partial c_i}{\partial t} = \gamma^c([c_i^h, \sum_{j\in\mathcal{N}(i)}(m_{ij}^{a} + m_{ij}^{d})]).
	\end{split}
\end{equation}

In Equation \ref{eq:ADBlock}, the advection term and the diffusion term are computed by $\phi^{a}$ and $\phi^{d}$, respectively, to generate the corresponding advection and diffusion messages.
The advection and diffusion messages are summed over the neighborhood of node $i$.
The update MLP $\gamma^{c}$, then predicts the time derivative of the scalar quantity.

\paragraph{Multiphysics Coupling}

\begin{figure}[t]
    \centering
    \includegraphics[width=0.9\columnwidth]{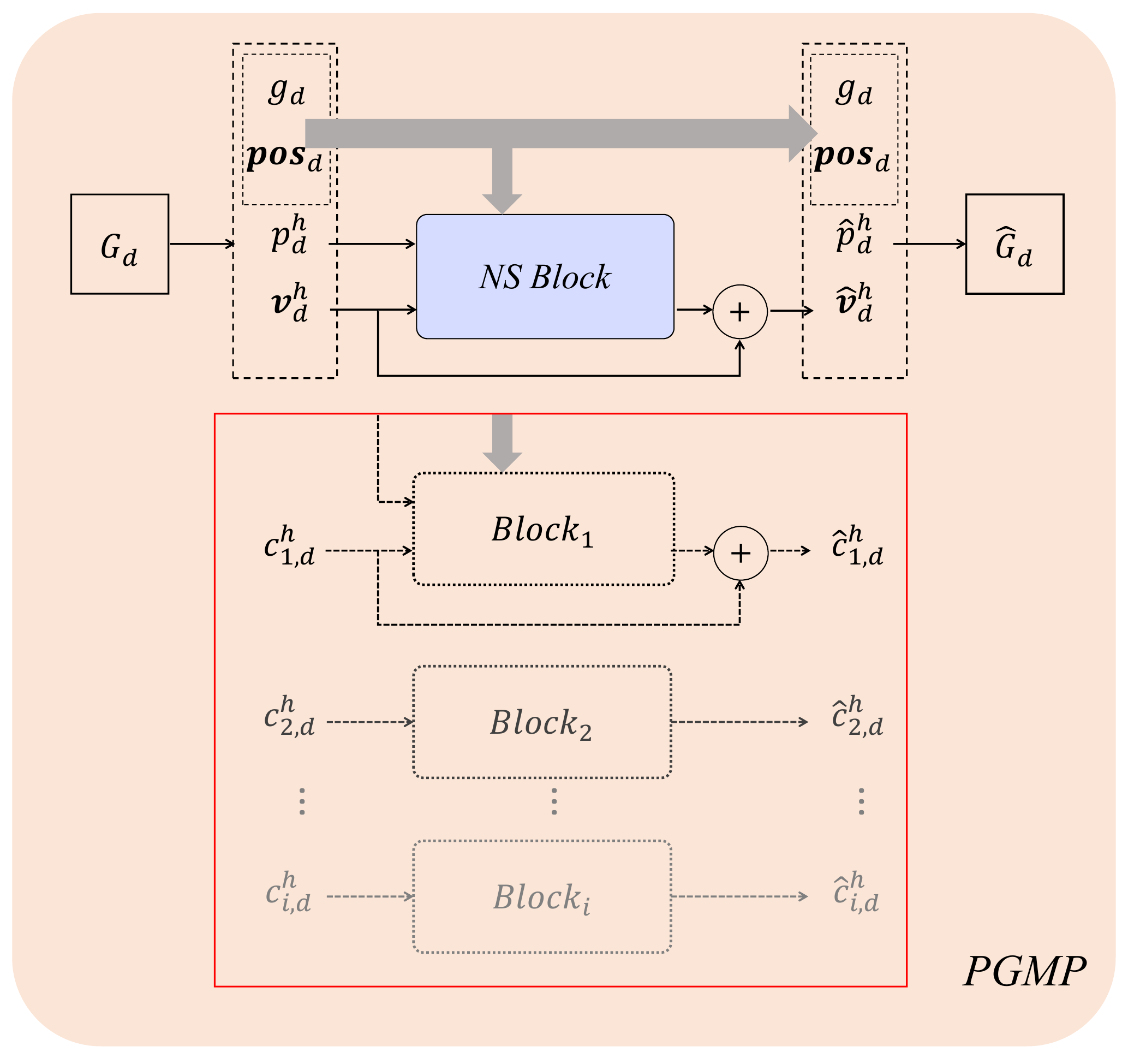}
    \caption{Multi-Task Adaptability of PGMP. The component within the red box can be omitted for single-phase flow, or extended to predict multiple scalar fields.}
    \label{fig:fig2}
\end{figure}

As shown in Figure \ref{fig:architecture}(b), the PGMP module incorporates a one-way coupling mechanism. 
Specifically, the NS Block directly predicts the updated pressure, while the time derivative it produces is integrated with the original velocity to obtain the updated velocity. This updated velocity serves not only as the output of the PGMP module but also as the input to the AD Block. The AD Block subsequently computes the evolution of the scalar field, which is also updated through time integration.
Therefore, the update of the fluid field is independent of the scalar field, relying solely on the fluid's internal dynamics. On the other hand, the evolution of the scalar field depends on the updated velocity field, directly reflecting the flow's driving influence on the transport process.
This coupling sequence adheres to the physical evolution logic of convection-diffusion problems in fluid mechanics and, through clearly defined interfaces between the blocks, offers the flexibility to extend to more complex multiphysics systems. For instance, the extensions could include multi-component transport and adding or ignoring scalar fields. 
As shown in Figure \ref{fig:fig2}, the PGMP module can be simplified by ignoring the scalar field $c$, enabling focused simulations on the fluid dynamics alone, while additional blocks can be added to predict more scalar fields.
The ability to decouple the concentration and fluid fields within the PGMP framework allows for a targeted approach to different simulation needs, making the system adaptable to varying degrees of complexity and different physical setups.

\begin{figure*}[t]
    \centering
    \includegraphics[width=0.95\linewidth]{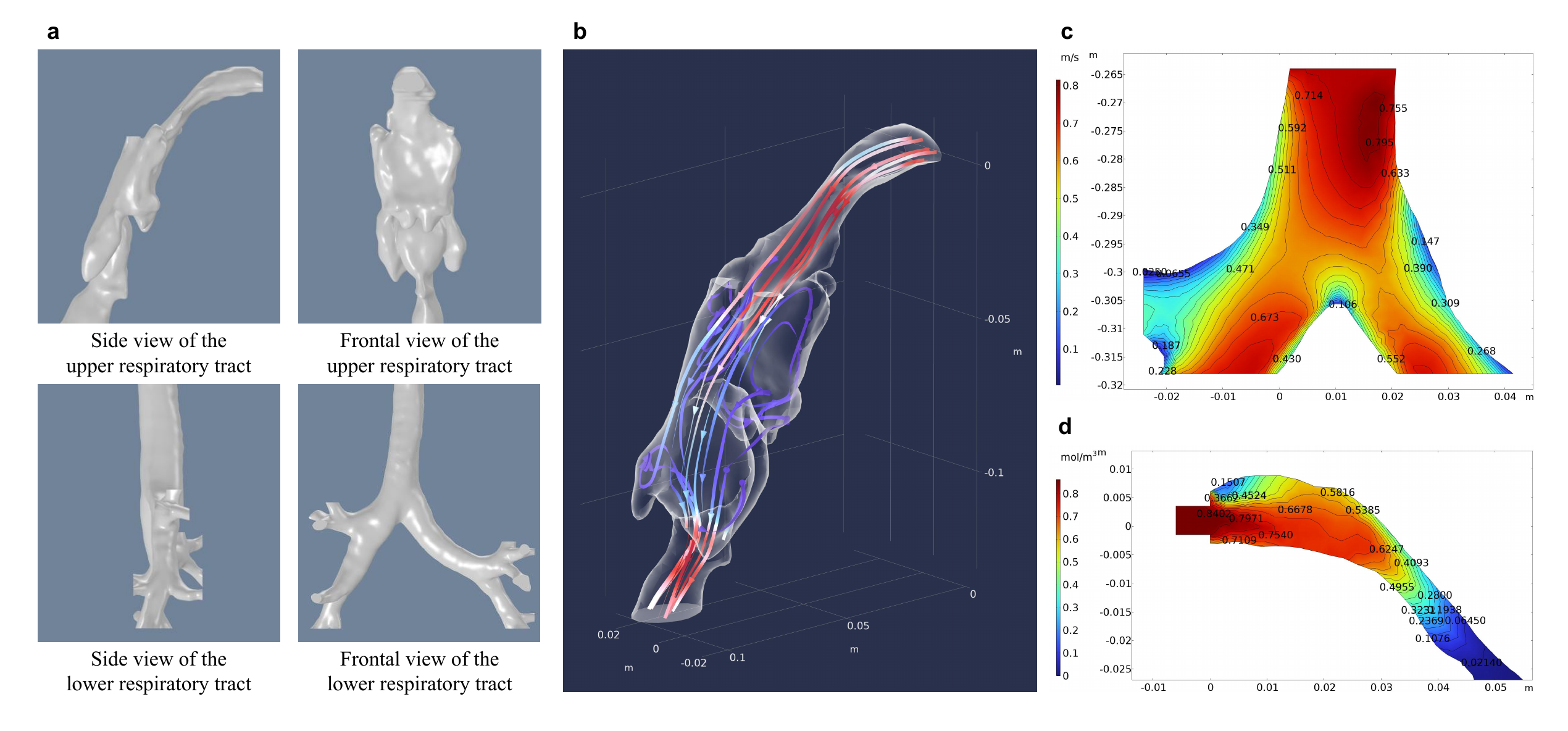}
    \caption{Overview of respiratory simulation datasets generated by traditional numerical methods: (a) Partial view of the human airway. (b) Streamlines of inhaled airflow in the upper respiratory tract, showing complex flow with backflow and vortices. (c) Cross-section at the lower tract bifurcation, colored by speed magnitude with contour lines. (d) Cross-section of the upper tract with a mouthpiece, colored by concentration magnitude with contour lines.}
	\label{fig:figdataset}
\end{figure*}

\subsection{Model Enhancements: Multi-Scale Design}

While PEGNet is built around the PDE-guided message passing approach, we integrate bi-stride multi-scale pooling \cite{cao2023efficient} to efficiently capture global and local dynamics. This non-learnable pooling reduces mesh complexity with increasing network depth, enabling fast computation on large geometries like human airways, and mitigates over-smoothing by employing skip connections, similar to U-Net \cite{ronneberger2015u}. 
However, unlike the one-MP (message passing) used in the Bi-Stride Multi-Scale Graph Neural Network \cite{cao2023efficient}, the PGMP module at each scale better preserves physical semantics.

\section{Experiment}
\label{sec:experiment}

\subsection{Experiment Setup}
\label{sec:setup}
\paragraph{Datasets}
From MeshGraphNets (MGN) \cite{pfaff2020learning}, we selected the publicly released dataset: cylinder flow, which is a fluid simulation dataset based on the incompressible Navier-Stokes equations.
Following this, we established respiratory simulation datasets, which comprise both airflow simulation and drug delivery simulation components.
To validate the PDE-guided message passing approach's applicability to other PDEs, we also constructed the Gray-Scott reaction-diffusion dataset.
The details of these datasets are included in Appendix A.

\begin{table*}[t]
	\caption{Comparison of different models in long-term prediction.}
	\centering
	\small
	\begin{tabular}{llccccc}
	\toprule
	\textbf{Case} & \textbf{Measurements} & \textbf{PEGNet(ours)} & \textbf{BSMS-GNN} & \textbf{MS-GNN-GRID} & \textbf{MGN} & \textbf{GraphUNets} \\
	\midrule
	\multirow{4}{*}{\makecell[l]{\vspace{1ex}Cylinder Flow\vspace{1ex}}}
	& RMSE-1 {[}1e-2{]}       & 0.216   & \textbf{0.204}    & 0.220   & 0.226  & 0.809\\
	& RMSE-50 {[}1e-2{]}        & \textbf{1.55}   & 2.42    & 2.74   & 4.39  & 18.7 \\
	& RMSE-last {[}1e-2{]}         & \textbf{2.56}   & 8.37    & 8.49   & 9.06  & 165 \\
	\midrule
	\multirow{4}{*}{\makecell[l]{\vspace{1ex}Airflow\vspace{1ex}}}
	& RMSE-1 {[}1e-2{]}       & \textbf{0.781}   & 0.797    & 0.790   & 0.883 & 12.1 \\
	& RMSE-50 {[}1e-2{]}        & \textbf{1.09}   & 2.40    & 1.93  & 8.35 & 26.7 \\ 
	& RMSE-last {[}1e-2{]}         & \textbf{8.30}   & 44.1   & 59.2  & 44.8 &  911\\
	\midrule
	\multirow{4}{*}{\makecell[l]{\vspace{1ex}Drug Delivery\vspace{1ex}}}
	& RMSE-1 {[}1e-3{]}       & 1.37   & \textbf{0.914}    & 1.21 & 2.05 & 28.2 \\
	& RMSE-50 {[}1e-3{]}        & \textbf{2.15}   & 2.47    & 5.32 & 6.08 & 30.6 \\
	& RMSE-last {[}1e-3{]}         & \textbf{5.19}   & 5.68    & 15.6 & 19.2 & 601 \\
	\bottomrule
	\end{tabular}
	\label{tab:RMSE}
\end{table*}

\begin{figure*}[t]
    \centering
    \includegraphics[width=0.85\textwidth]{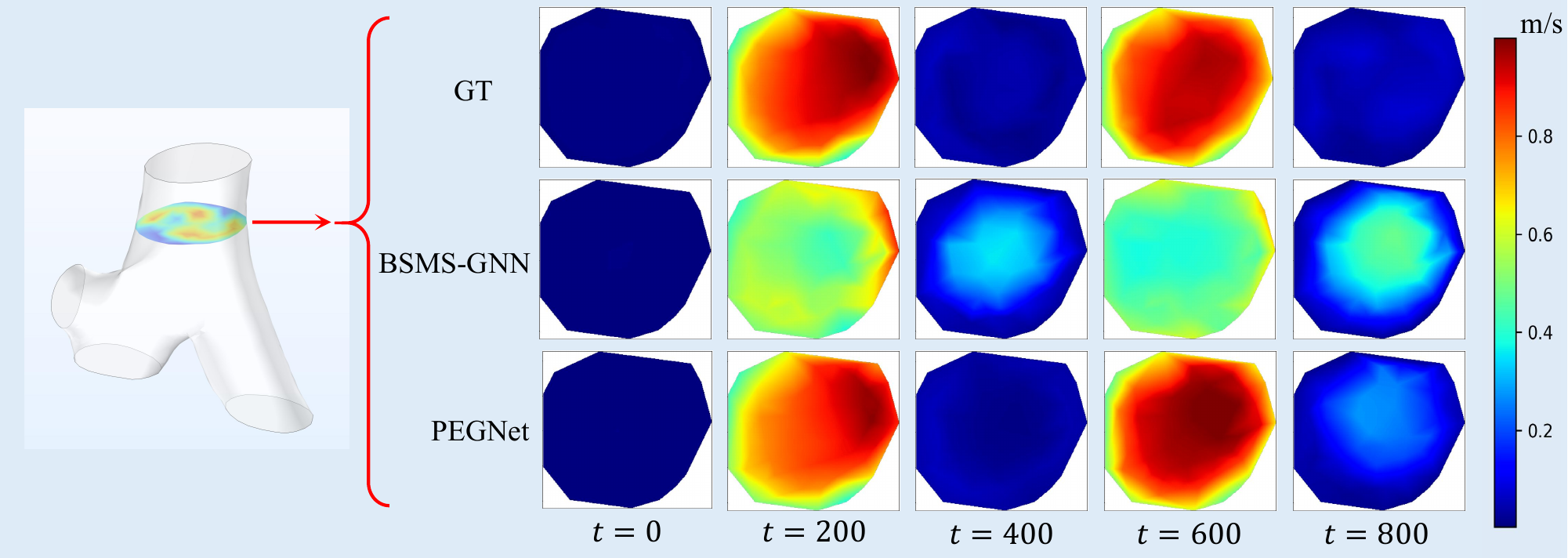}
    \caption{From $t=0$ to $t=800$ represents a complete inhalation-exhalation cycle, with key points at $t=0$ (start of inhalation), $t=200$ (peak inhalation velocity), $t=400$ (start of exhalation), $t=600$ (peak exhalation velocity), and $t=800$ (end of exhalation). The BSMS-GNN model shows increasing artifacts in the central airway, predicting airflow even after exhalation ends, contrary to reality. In contrast, our model remains more stable with only minor artifacts near the cycle's end.}
    \label{fig:airflow}
\end{figure*}

\paragraph{Model Training}
In this work, following the principles of PINNs \cite{raissi2019physics}, we incorporate physical constraints into the training process, directly guiding the model output to adhere to the underlying physical laws while retaining the efficiency and flexibility of data-driven learning. The loss function is designed to include the following terms:
(1) \textit{Node Prediction Error} ($L_{\text{pred}}$): This term is computed as the mean squared error (MSE) between the predicted node feature $\hat{p}_i$ and the true features $p_i$ at each node, which is expressed as:
\begin{equation}
    L_{\text{pred}} = \frac{1}{N} \sum_{i=1}^{N} \left( \hat{p}_i - p_i \right)^2.
\end{equation}
(2) \textit{Divergence Regularization Loss} ($L_{\text{div}}$): This term enforces the divergence-free condition on the velocity field, which is defined as:
\begin{equation}
	\label{eq:Ldiv}
    L_{\text{div}} = \frac{1}{N} \sum_{i=1}^{N} \left( \sum_{j \in \mathcal{N}(i)} \frac{1}{|\mathcal{N}(i)|} \left( \bm{v}_j - \bm{v}_i \right) \cdot \bm{d}_{ij} \right)^2.
\end{equation}
(3) \textit{Mass Conservation Regularization Loss} ($L_{\text{mass}}$): This term ensures compliance with mass conservation principles, which is defined as:
\begin{equation}
	\label{eq:Lmass}
    \begin{split}
        L_{\text{mass}} = &\frac{1}{N} \sum_{i=1}^{N} \bigg( \left( c_i^{t+1} - c_i^t \right) + \\
        & \sum_{j \in \mathcal{N}(i)} \left( (\bm{v}_i^{t+1} \cdot \bm{d}_{ij}) c_i^t - (\bm{v}_j^{t+1} \cdot \bm{d}_{ij}) c_j^t \right) \bigg)^2.
    \end{split}
\end{equation}
The specific definitions for these formulas are provided in the Appendix B. Thus, the total loss function can be expressed as:
\begin{equation}
L_{\text{total}} = L_{\text{pred}} + \lambda_{\text{div}} L_{\text{div}} + \lambda_{\text{mass}} L_{\text{mass}},
\end{equation}
where $\lambda_{\text{div}}$ and $\lambda_{\text{mass}}$ are hyperparameters controlling the respective regularizations. 

For optimization, we used the AdamW optimizer \cite{loshchilov2019decoupledweightdecayregularization}, which incorporates decoupled weight decay and adaptive moment estimation, along with a warmup cosine decay learning rate scheduler \cite{vaswani2017attention}. 

\paragraph{Baseline}
We compare our model with several representative baseline models, including MGN \cite{pfaff2020learning}, BSMS-GNN \cite{cao2023efficient}, MS-GNN-GRID \cite{lino2022multi}, and GraphUNets \cite{gao2019graph}. 
A detailed description of these models is provided in Appendix C.

\paragraph{Evaluation Metrics}

The model performance is evaluated using the following metrics: (1) \textit{Root Mean Squared Error (RMSE)}, which measures the accuracy of node predictions and is defined as: 
$\textit{RMSE} = \sqrt{\frac{1}{N} \sum_{i=1}^{N} \| \hat{p}_i - p_i \|^2 }$;
(2) \textit{Divergence Error (DVE)}, defined as the square root of the divergence regularization loss, which assesses the divergence-free condition of the velocity field:
$\textit{DVE} = \sqrt{L_{\text{div}}}$;
(3) \textit{Mass Conservation Error (MCE)}, defined as the square root of the mass conservation regularization loss, which evaluates the consistency of mass transport across nodes:
$\textit{MCE} = \sqrt{L_{\text{mass}}}$.
Lower values of these metrics indicate better accuracy and improved physical consistency.
Detailed explanations of these metrics are provided in Appendix B.

\begin{table}[t]
	\caption{Multi-channel evaluation of velocity (Vel), pressure (Pre), and concentration (Conc) using average RMSE[1e-2] over all timesteps.}
	\centering
    \small

	\begin{tabular}{c c c c c c}
	\toprule
	\makecell[l]{\textbf{Method}} & \multicolumn{2}{c}{\textbf{Airflow}} & \multicolumn{3}{c}{\textbf{Drug Delivery}}\\
	\cmidrule(lr){2-3} \cmidrule(lr){4-6}
	& Vel & Pre & Vel & Pre & Conc\\
	\midrule
	\makecell[l]{\textbf{MGN}} & 20.5  & 52.3 & 0.823 & 2.15 & 2.32 \\
	\makecell[l]{\textbf{BSMS-GNN}} & 14.3  & 68.6 & 0.146 & 0.0522 & 1.28 \\
	\makecell[l]{\textbf{PEGNet(ours)}} & \textbf{4.67} & \textbf{3.87} & \textbf{0.145} & \textbf{0.0141} & \textbf{0.648}\\
	
	\bottomrule
	\end{tabular}
	\label{tab:consistency}
\end{table}

\subsection{Benchmarking and Evaluation}

\paragraph{Cylinder Flow} 
In Table \ref{tab:RMSE}, for the Case Cylinder Flow, the data for the baseline models comes from BSMS-GNN \cite{cao2023efficient}.
To ensure experimental fairness, PEGNet's training hyperparameters were selected to closely match those of BSMS-GNN \cite{cao2023efficient}, as detailed in Appendix B. 
After training the model to convergence on the training set, we evaluated its performance on the test set using an autoregressive rollout to generate trajectories of the same length as the ground truth.
We then computed the RMSE at time steps 1, 50, and the last step across all prediction trajectories and averaged the results at each step.
The results are shown in the Case Cylinder Flow of Table \ref{tab:RMSE}. The bolded values in the table represent the best results within the same measurement. 
At the first step, most models perform well. However, in the subsequent time steps, our model consistently maintains a significant advantage, demonstrating its stronger long-term predictive stability.

\paragraph{Airflow}

After the preliminary evaluation of PEGNet's effectiveness in the 2D fluid prediction task, we further assessed its performance in simulating airflow in the 3D human airway.
All models were trained to convergence on the training set, and the same evaluation strategy as in the Cylinder Flow case was adopted.
The corresponding results are shown in the Case Airflow of Table \ref{tab:RMSE}, which demonstrate that PEGNet outperforms baseline models at time steps 1, 50, and the last step. 
Figure \ref{fig:airflow} compares the flow velocities for the ground truth (GT), PEGNet, and BSMS-GNN at the airway cross-section. As shown, PEGNet's predictions align more closely with the ground truth and exhibit greater stability across all time steps.
Additional visualizations of the model predictions are included in Appendix E.

\begin{table}[t]
	\caption{Evaluation using DVE{[}1e-6{]} and MCE{[}1e-5{]}.}
	\centering
	\small
    \setlength{\tabcolsep}{1mm}

	\begin{tabular}{llccc}
	\toprule
	\textbf{Case} & \textbf{Metrics} & \textbf{PEGNet(ours)} & \textbf{BSMS-GNN} & \textbf{MGN}  \\
	\midrule
	\multirow{3}{*}{\makecell[l]{Airflow}}
	& DVE-1       & \textbf{0.647}   & 0.685  & 1.25 \\
	& DVE-50         & \textbf{11.1}   & 11.4    & 12.7 \\
	& DVE-last          & \textbf{35.1}   & 250    & 268 \\
	\midrule
	\multirow{3}{*}{\makecell[l]{Drug Delivery}}
	& DVE-1        & \textbf{0.218}   & 0.416    & 0.921 \\
	& DVE-50        & \textbf{2.39}   & 2.46    & 5.37 \\
	& DVE-last          & \textbf{1.13}   & 1.17    & 8.80 \\
	\midrule
	\multirow{3}{*}{\makecell[l]{Drug Delivery}}
	& MCE-1        &  \textbf{27.9}  &  38.5   &  278 \\
	& MCE-50         &  \textbf{5.95}  &  6.40    & 17.4 \\
	& MCE-last          &  \textbf{7.18}  &  7.82   & 18.5 \\
	\bottomrule
	\end{tabular}
	
	\label{tab:DVEMCE}
\end{table}

\paragraph{Drug Delivery}

Since PEGNet demonstrated the ability to simulate airflow in the 3D human airway, we further applied it to the multiphysics drug delivery prediction task. The evaluation procedure remained the same as in the previous two cases.
As shown in the Case Drug Delivery of Table \ref{tab:RMSE}, the results indicate that although PEGNet does not show a significant advantage in first-step prediction, it exhibits superior stability compared to other models in subsequent time steps.

\paragraph{Physical Consistency} 
Table \ref{tab:consistency} presents a multi-channel comparison between PEGNet and two representative models. The results show that, although MGN and BSMS-GNN still exhibit a certain degree of accuracy in overall predictions or velocity field predictions, their performance in predicting coupled physical fields is significantly worse than that of PEGNet, whether coupled with the pressure field or further extended to the concentration field.
This indicates that MGN and BSMS-GNN have limitations in learning physical principles governing each quantity and in handling multiphysics coupling. In contrast, PEGNet incorporates physical prior knowledge and is better at capturing the relationships between different physical quantities.
Table \ref{tab:DVEMCE} provides another comparison of these models, using the physical consistency metrics DVE and MCE.
We computed DVE and MCE at time steps 1, 50, and the last step across all generated trajectories and averaged the values at each corresponding step, which highlights that our model better adheres to physical constraints over long rollouts than the others.

\paragraph{Gray-Scott Reaction-Diffusion}
To further validate the generality of our PDE-guided message passing approach beyond fluid dynamics, we extended it to the Gray-Scott equation, with details provided in Appendix D. The results demonstrate its potential applicability to other systems governed by PDEs.

\subsection{Ablation Study}
To evaluate the contributions of the PGMP module and the physics-informed loss, we conducted ablation experiments on the airflow dataset by progressively removing these components. 
Four models were tested: the full PEGNet (Ours); PEGNet without the physics-informed loss (Model A); PEGNet without the PGMP module, using standard message passing instead (Model B); and PEGNet without both modules (Model C). 
Results in Table \ref{tab:ablation} show that removing both modules (Model C) yields the worst long-term predictions, while including either significantly improves accuracy and stability. The complete PEGNet achieves the best performance in long-term rollouts, demonstrating the synergistic benefits of both components.

\begin{table}[t]
	\centering
    \small
    \setlength{\tabcolsep}{1mm}
	\caption{Results of the Ablation Study.}
	\label{tab:ablation}
	\begin{tabular}[width=0.5\columnwidth]{lcccc}
	\toprule
	\textbf{Measurements} & \textbf{Ours} & \textbf{Model A} &  \textbf{Model B} &  \textbf{Model C} \\
	\midrule
	
	\multirow{1}{*}{\makecell[l]{RMSE-1 {[}1e-2{]}}}
	& {0.781}  & \textbf{0.757} & 0.788 & 0.797\\

	\multirow{1}{*}{\makecell[l]{RMSE-50 {[}1e-2{]}}}
	& \textbf{1.09}  & 1.26 & 1.54 & 2.40\\

	\multirow{1}{*}{\makecell[l]{RMSE-last {[}1e-2{]}}}
	& \textbf{8.30}  & 12.7 & 30.8 & 41.1\\
	\bottomrule
	\end{tabular}
	
\end{table}

\section{Conclusion}
\label{sec:conclusion}

This study proposes PEGNet, a Physics-Embedded Graph Network that employs the PDE-guided message passing approach for stable modeling of airflow and drug delivery in the human airway. 
PEGNet implements this approach through a Physics-Guided Message Passing (PGMP) module, which is specifically based on the incompressible Navier-Stokes equations and the Advection-Diffusion equation.
Overall, our approach enhances the neural model's long-term predictive accuracy and physical consistency. The results demonstrate that it outperforms other data-driven methods in both aspects.

Despite the promising results, several limitations remain. These include reliance on simplified airway models, limited data availability, and superficial integration of physical principles, all of which constrain the model's applicability to real-world scenarios.
Additionally, the Eulerian framework used for drug delivery struggles to accurately capture particle trajectories.
Future work will focus on incorporating more anatomically realistic and variable 3D airway models, expanding data sources beyond simulations, and strengthening the integration of physical principles with neural networks.
In particular, exploring Lagrangian-based formulations and accounting for more complex pulmonary physiological conditions will be key to improving both model performance and clinical applicability.



\section*{Acknowledgment}

This work was supported in part by the National Natural Science Foundation of China under Grant No. 52535009 and No. 62276222; in part by the Fundamental Research Funds for the Central Universities under Grant No. 20720250164; in part by the Xiamen Natural Science Foundation under Grant No. 3502Z202571027.

\bibliographystyle{IEEEtran}
\bibliography{references}

\appendices

\setcounter{figure}{0}
\setcounter{table}{0}
\setcounter{equation}{0} 

\renewcommand{\theequation}{\thesection.\arabic{equation}}
\renewcommand{\thefigure}{\thesection.\arabic{figure}}
\renewcommand{\thetable}{\thesection.\arabic{table}}

\section{Dataset Details}
\label{ap:datasets}

\begin{figure*}
    \centering
    \includegraphics[width=1\textwidth]{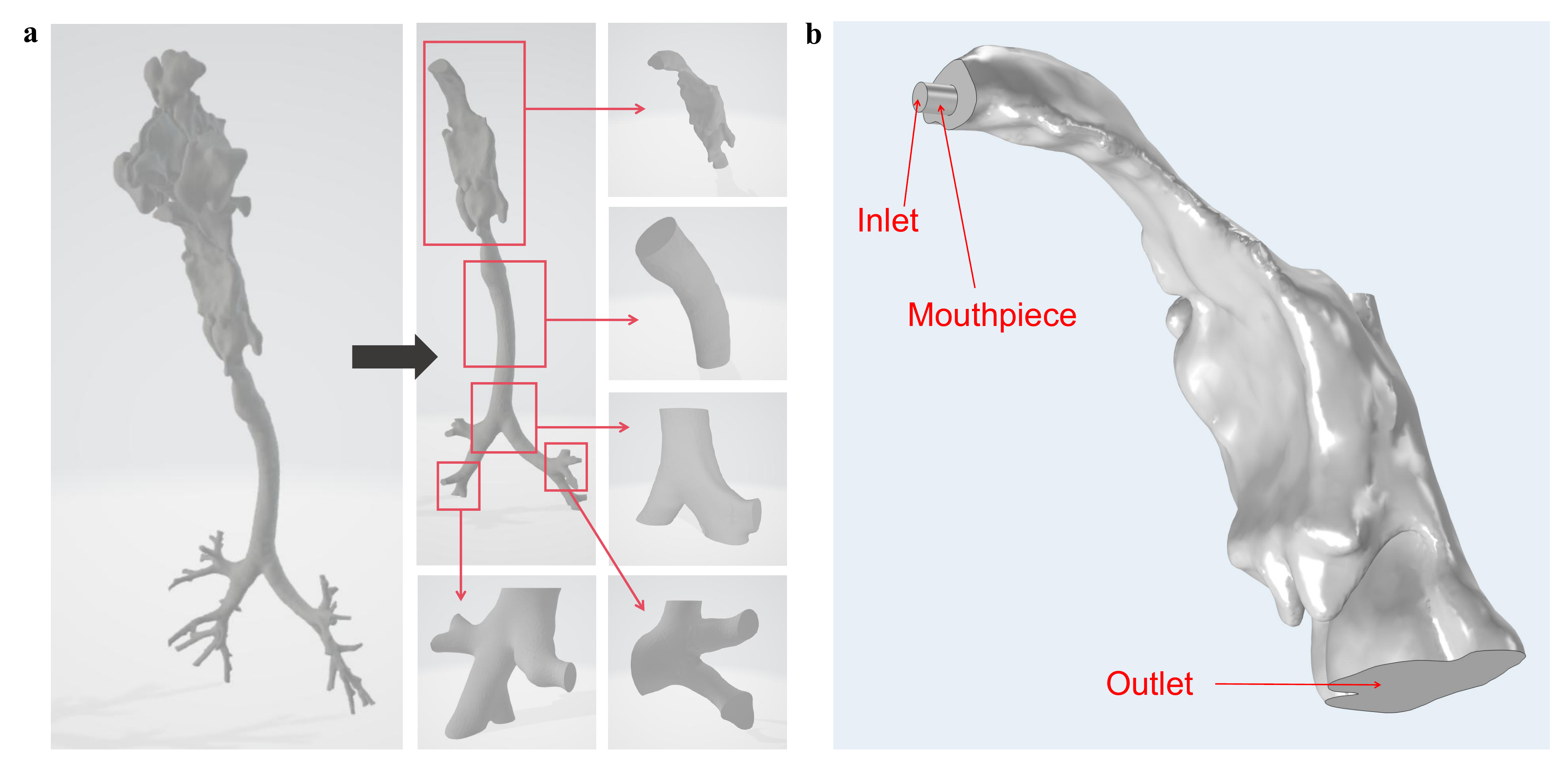}
    \caption{(a) Simplification of human airways, which are used for the airflow simulation. (b) Upper airway model for drug delivery simulation. Unlike the model used in the airflow simulation, this simulation adds a mouthpiece in front of the oral cavity  to simulate the entry of drugs and airflow into the mouth via a specific device.}
	\label{fig:airway}
\end{figure*}



\subsection{Respiratory Simulation Datasets} 

\paragraph{Airflow Simulation Dataset.}
The airflow simulation is conducted based on 3D airway models provided by \cite{bartol2024computational}. As shown in Figure \ref{fig:airway}(a), we pruned and segmented the airway model into five distinct parts, each representing a major section of the respiratory system. 
This simplification ensures that the simulations capture the essential flow dynamics while maintaining computational feasibility. 
The final dataset includes 120 trajectories, each reflecting variations in geometry or boundary flow conditions. Among them,  100 were used for training and 20 for testing. The boundary conditions were strictly based on existing research \cite{elcner2016numerical}. 
Each time step was captured with a resolution of 0.005 seconds, with a total of 800 time steps per trajectory, yielding a complete inhalation and exhalation cycle lasting 4 seconds.



\paragraph{Drug Delivery Simulation Dataset.} 

Building upon the airflow simulation, this study further incorporates an Eulerian method \cite{pourhashem2020eulerian} to model the drug concentration field. This approach simplifies the representation of drug particles by focusing solely on their spatial concentration, disregarding their shape and size. 
In doing so, the method reduces drug delivery to an advection-diffusion process governing the concentration field.
However, it still maintains sufficient complexity to effectively evaluate the GNN model's capability in handling multi-physics simulations.
Given the crucial role of the upper airway in the initial transport of drug particles, the current simulation specifically focuses on drug delivery within the upper respiratory tract, as illustrated in Figure~\ref{fig:airway}(b).
This final dataset includes 144 trajectories under varying airflow and drug concentration conditions, with 120 used for training and 24 for testing.
The boundary conditions were specified in accordance with the settings established by \cite{pourhashem2020eulerian}.
For each trajectory, a time step resolution of 0.005 seconds is maintained, with a total of 400 time steps, yielding an inhalation phase lasting 2 seconds. 


\subsection{Gray-Scott Reaction-Diffusion Dataset}
This dataset is generated based on the Gray-Scott Reaction-Diffusion equations, defined as follows:

\begin{figure}[t]
    \centering
    \includegraphics[width=0.7\columnwidth]{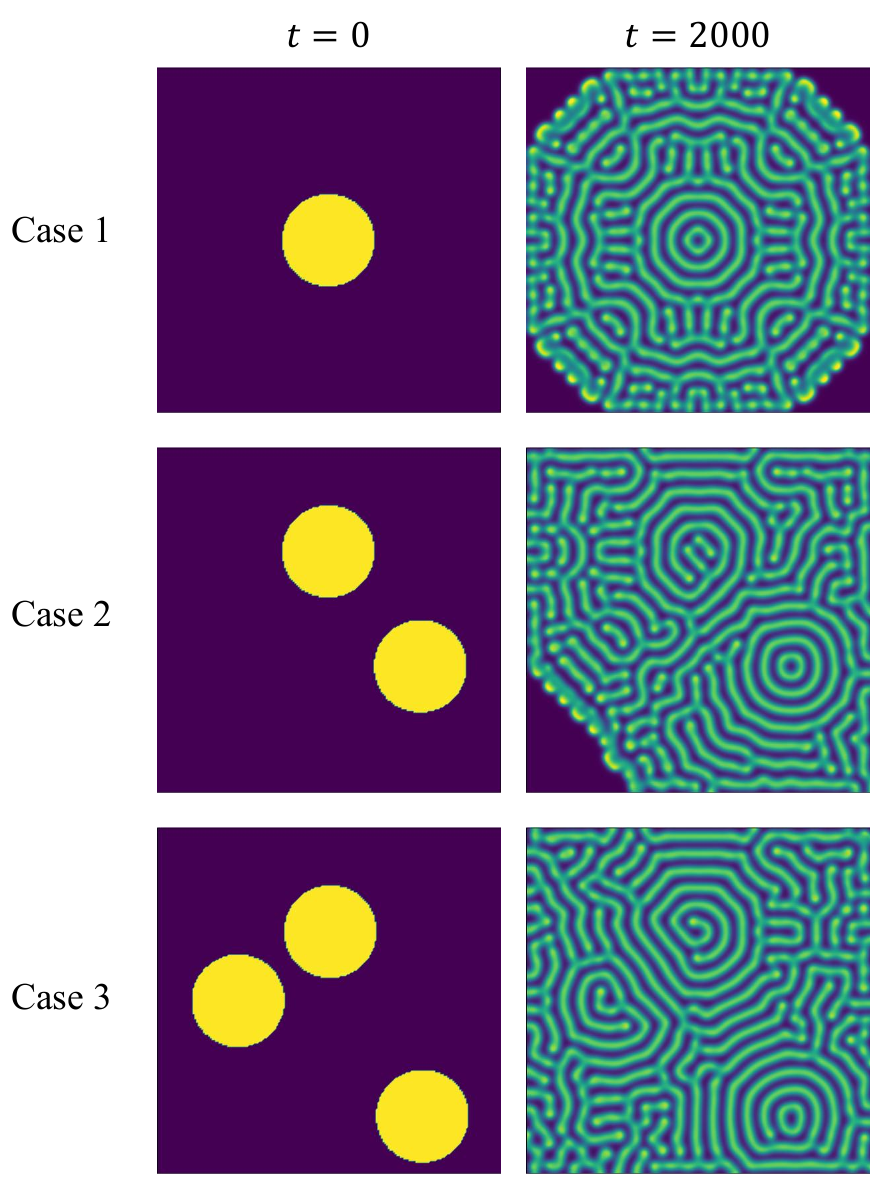}
    \caption{Different initial and final states of Gray-Scott reaction-diffusion simulations. The figure shows the concentration distribution of $u$. 
	In the initial state, the yellow circular region represents the concentrated distribution area of $u$, while $v$ is evenly spread across the entire plane. 
	In the final state, the concentration distribution of u stabilizes into complex patterns.}
	\label{fig:GSdataset}
\end{figure}

\begin{equation}
	\label{eq:GS}
	\begin{split}
	\frac{\partial u}{\partial t} &= D_u \nabla^2 u - uv^2 + F(1 - u), \\
	\frac{\partial v}{\partial t} &= D_v \nabla^2 v + uv^2 - (F + k)v,
	\end{split}
\end{equation}
where $u(x,y,t)$ and $v(x,y,t)$ denote the concentrations of two chemical species. $D_u$ and $D_v$ are their respective diffusion coefficients, $F$ is the external feed rate of $u$, and $k$ is the removal rate of $v$.
The Gray-Scott reaction-diffusion model is primarily used to simulate complex pattern formation in biological organisms, autocatalytic chemical reactions, and the formation and evolution of material microstructures.
In our simulation, $D_u$ and $D_v$ are 2e-5 and 0.5e-5 respectively, $F$ is 0.055, and $k$ is 0.062.
The entire experiment consists of 24 trajectories, with 16 used for training and 8 for testing. Each trajectory comprises 2000 time steps, with a resolution of one second per step. 
Different initial concentration distributions across trajectories lead to varying final concentration patterns, as shown in Figure \ref{fig:GSdataset}.

\begin{table*}
	\caption{Details of the data included in the datasets. }
	\centering

	\begin{tabular}{llccc}
	\toprule
	\textbf{Case} & \textbf{Mesh Type} & \textbf{Data Fields} & \textbf{Steps} & \textbf{$\Delta t$}  \\
	\midrule
	Cylinder Flow & triangle 2D \vspace{1pt} & node type, cells, mesh pos, velocity, pressure & 600& 0.01\\

	Airflow & tetrahedron 3D \vspace{1pt} & node type, cells, mesh pos, velocity, pressure & 800& 0.005 \\               				

	Drug Delivery & tetrahedron 3D & node type, cells, mesh pos, velocity, pressure, concentration & 400 & 0.005 \\

	Gray-Scott & triangle 2D & node type, cells, mesh pos,cu,cv & 2000& 1 \\
	\bottomrule
	\end{tabular}
	
	\label{tab:datasets}
\end{table*}

\begin{table*}[t]
	\caption{Model input and output settings. }
	\centering

	\begin{tabular}{lcccc}
	\toprule
	\textbf{Case} & \textbf{Spatial Dimension} & \textbf{Inputs}& \textbf{Outputs} & \textbf{Output dimension}\\
	\midrule
	Cylinder Flow & 2 & $\bm{v},p,pos,g,n$ & $\dot{\bm{v}},p$ & 3 (2+1)\\
	
	Airflow & 3 & $\bm{v},p,pos,g,n$ & $\dot{\bm{v}},p$ & 4 (3+1)\\
	
	Drug Delivery& 3 &$\bm{v},c,p,pos,g,n$ & $\dot{\bm{v}},\dot{c},p$ & 5 (3+1+1)\\

	Gray-Scott& 2 & $u,v,pos,g,n$ & $\dot{u},\dot{v}$ & 2 (1+1)\\

	\bottomrule
	\end{tabular}
	
	\label{tab:inputoutput}
\end{table*}

\section{Experiment Details}
\label{ap:training}

\subsection{Implementation Details}

All experiments were conducted on a single Intel Xeon Gold 5222 CPU and a single Nvidia A40 GPU.
The model was implemented with PyTorch and PyTorch Geometric (PyG) \cite{fey2019fastgraphrepresentationlearning}.
To ensure fairness with the baseline model in the Cylinder Flow case, we adopted the following configuration, which is also used in other experiments.
The MLPs within our model consist of three ReLU-activated hidden layers. Both hidden and output layers are sized at 64, except for the decoder, which matches the target tensor's output size. Layer Normalization is applied to all MLP outputs, except those of the Decoder. The multi-level architecture of the model has a depth of 5.
Furthermore, as shown in Table 2, different tasks are configured with different inputs and outputs, requiring certain modifications to the model's Encoder and Decoder.
In addition to physical quantities, the input also includes grid node positions $pos$, mesh edge connectivity information $g$, and node types $n$. 
Following the rules of MGN \cite{pfaff2020learning}, the node types are encoded as one-hot vectors which are then concatenated with the physical quantities to form the complete input.

We used the AdamW optimizer with a peak learning rate of 1e-4 and a weight decay of 1e-4. A cosine decay scheduler with 20,000 warmup steps was applied. The batch size was set to 4.
All experiments were trained until convergence, rather than for a fixed number of iterations. We considered training to have converged when the mean squared error (MSE) on the validation set ceased to exhibit significant improvement.

\subsection{Detailed Definition of the Loss Function}
Following MGN \cite{pfaff2020learning}, we employ a self-supervised method. We randomly sample node state pairs ($u_i^{t}$, $u_i^{t+1}$) from training trajectories. We use $u_i^{t}$ to compute each node's feature $\hat{u_i}^{t+1}$, and then train our model by minimizing the MSE between the model's prediction and the corresponding ground truth $u_i^{t+1}$.
In addition to the data-driven objective, we incorporated physical consistency into the training process. Specifically, two physics-based regularization terms were added to the loss function, tailored to the final task of pulmonary drug delivery.

\paragraph{Divergence-Free Condition.} For the cylinder flow, airflow simulation, and drug delivery simulation datasets used in this paper, fluid flow is modeled using the incompressible Navier-Stokes equations. As a result, the fluid velocity consistently satisfies the divergence-free condition, expressed by the equation below:

\begin{equation}
	\quad \nabla \cdot \bm{v} = 0.
\end{equation}
After discretizing at each node of the mesh, we obtain the following regularization term:
\begin{equation}
	\label{eq:divergence at nodes}
	L_{\text{div}} = \frac{1}{N} \sum_{i=1}^{N} (\nabla \cdot \bm{v}_i)^2,
\end{equation}
where the divergence of the velocity field $\nabla \cdot \bm{v}_i$ can be numerically approximated using differences with neighboring nodes.
We define $\Delta \bm{x}_{ij} = \bm{x}_i-\bm{x}_j$ as the position vector difference between node pairs, and $\bm{d}_{ij}=\frac{\Delta \bm{x}_{ij}}{\|\Delta \bm{x}_{ij}\|}$ as the corresponding normalized direction vector. The divergence at node $i$ can then be approximated by:

\begin{equation}
	\bm{v}_i \approx \sum_{j \in \mathcal{N}(i)} w_{ij} \left( \left( \bm{v}_j - \bm{v}_i\right) \cdot \bm{d}_{ij}\right),
\end{equation}
where $w_{ij}$ is a weighting coefficient related to edge length or local geometry, used for numerical stability and scale consistency adjustment. In practical implementation, a simple uniform weight $w_{ij}=\frac{1}{\|\mathcal{N}(i)\|}$ is employed, where a is the number of neighbors of node $i$.
Based on the above, by substituting all terms into Equation \ref{eq:divergence at nodes}, the divergence regularization loss can be completely represented as:
\begin{equation}
    L_{\text{div}} = \frac{1}{N} \sum_{i=1}^{N} \left( \sum_{j \in \mathcal{N}(i)} \frac{1}{|\mathcal{N}(i)|} \left( \bm{v}_j - \bm{v}_i \right) \cdot \bm{d}_{ij} \right)^2,
\end{equation}

\begin{table}[t]
	\centering
	\caption{Regulation terms calculated using the ground truth.}
	\label{tab:GtDVEMCE}
	\begin{tabular}[width=0.8\columnwidth]{lccc}
	\toprule
	\textbf{Regularization Terms} & \textbf{Cylinder Flow} & \textbf{Airflow} & \textbf{Drug Delivery} \\
	\midrule
	
	\multirow{1}{*}{Divergence-Free($L_{\text{div}}$)}
	& 2.5e-7 & 1.28e-10 & 1.84e-11 \\

	\multirow{1}{*}{Mass conservation($L_{\text{mass}}$)}
	& -& - & 5.95e-9 \\  

	\bottomrule
	\end{tabular}
	
\end{table}

\paragraph{Mass Conservation Condition.} 
Drug delivery simulations involve more than just fluid flow. We've also brought in an additional physical field, the concentration field. 
In this context, we focus on local mass conservation, which means each node's concentration changes in direct response to the difference between its inflow and outflow.
Therefore, the regularization term can be expressed as:
\begin{equation}
	\label{eq: mass at nodes}
	L_{\text{mass}} = \frac{1}{N}\sum_{i=1}^{N}(\Delta c_i + {flux}_i)^2,
\end{equation}
where $\Delta c_i = c_i^{t+1} - c_i^{t}$ represents the change in concentration at node $i$ and ${flux}_i$ is the net outflow from node $i$. Ideally, these two terms are opposite in sign. We define ${flux}_i$ as:
\begin{equation}
	{flux}_i = \sum_{j in \mathcal{N}(i)}(f_{ij}-f_{ji}),
\end{equation}
where $f_{ij}$ is the drug flow from i to j, approximated as:
\begin{equation}
	f_{ij} \approx (\bm{v}_i \cdot \bm{d}_{ij})c_i.
\end{equation}
By substituting these terms into Equation \ref{eq: mass at nodes}, we can ultimately derive the full mass conservation regularization loss, as follows:
\begin{equation}
    \begin{split}
        L_{\text{mass}} = &\frac{1}{N} \sum_{i=1}^{N} \bigg( \left( c_i^{t+1} - c_i^t \right) + \\
        & \sum_{j \in \mathcal{N}(i)} \left( (\bm{v}_i^{t+1} \cdot \bm{d}_{ij}) c_i^t - (\bm{v}_j^{t+1} \cdot \bm{d}_{ij}) c_j^t \right) \bigg)^2,
    \end{split}
\end{equation}

\paragraph{The Reasonableness of These Regularization Terms.}
It's important to note that while both of the above regularization terms encourage stronger physical consistency as they approach zero, they may not reach zero even on the ground truth (GT) due to simplifications and approximations in their formulas. Therefore, these two terms are used as soft constraints. To test their reasonableness, we evaluated them on the ground truth.
The results are shown in Table \ref{tab:GtDVEMCE}. 
As can be seen, both regularization terms are very close to zero on the GT values, allowing them to be included as soft constraints in the loss function.

\paragraph{Evaluation Metrics.} 
Our evaluation metrics include the RMSE between the model's output and the corresponding ground truth, as well as the square roots of these two regularization terms. According to the preceding description, a smaller value for these metrics indicates better model performance. 
The square root is used to make the metrics more intuitively comparable, as it returns the value to the original physical units of the predicted quantity. This makes the error values directly interpretable and comparable to the actual values.

\section{Baseline Model Details}

We selected the best-performing hyperparameter settings from the experimental configurations of the BSMS-GNN \cite{cao2023efficient} and MGN \cite{pfaff2020learning} papers to construct the baseline model.

\paragraph{MeshGraphNets.} 
This model uses the encode-process-decode architecture with a processor of stacked message passing layers. In our experiments, we set the number of message passing layers to 15. Its MLPs are identical to ours, but with a hidden size of 128. We trained it with a batch size of 2 on each dataset until convergence, just like our model.

\paragraph{BSMS-GNN.}
With non-learnable downsampling and a multi-level architecture, this model offers high accuracy while staying lightweight. Following the original paper, we set the multi-level depth to 7 in our experiment. Its MLPs of message passing layers match ours but have a dimension of 128, and the batch size is 4.

\paragraph{MS-GNN-GRID.}
The combination of a multi-scale architecture and rotation equivariance enables it to more effectively learn the underlying physical laws and infer the temporal evolution of fluid flow.
Because of the unique downsampling algorithm, we set the initial grid cell size in our implementation to the average edge length of the original mesh. Then, the grid spacing is doubled at each level, resulting in a four-level hierarchical structure. The MLP configurations are kept the same as in the above models, with four-layer MLPs used in the first and fourth levels, and two-layer MLPs used in the second and third levels.
During training, a batch size of 2 was used.

\paragraph{GraphUNets.}
The learnable downsampling strategy leads to significant computational overhead. Therefore, we made several modifications, including replacing the GCN layers with the message passing layers used in BSMS and adopting a sparsified graph representation instead of the adjacency matrix. Finally, we adopted a seven-level hierarchical structure with a batch size set to 1.
\label{ap:baseline}

\section{Additional Experiments: Gray-Scott Reaction-Diffusion}
\label{ap:further experiment}

\begin{figure}
    \centering
    \includegraphics[width=1\columnwidth]{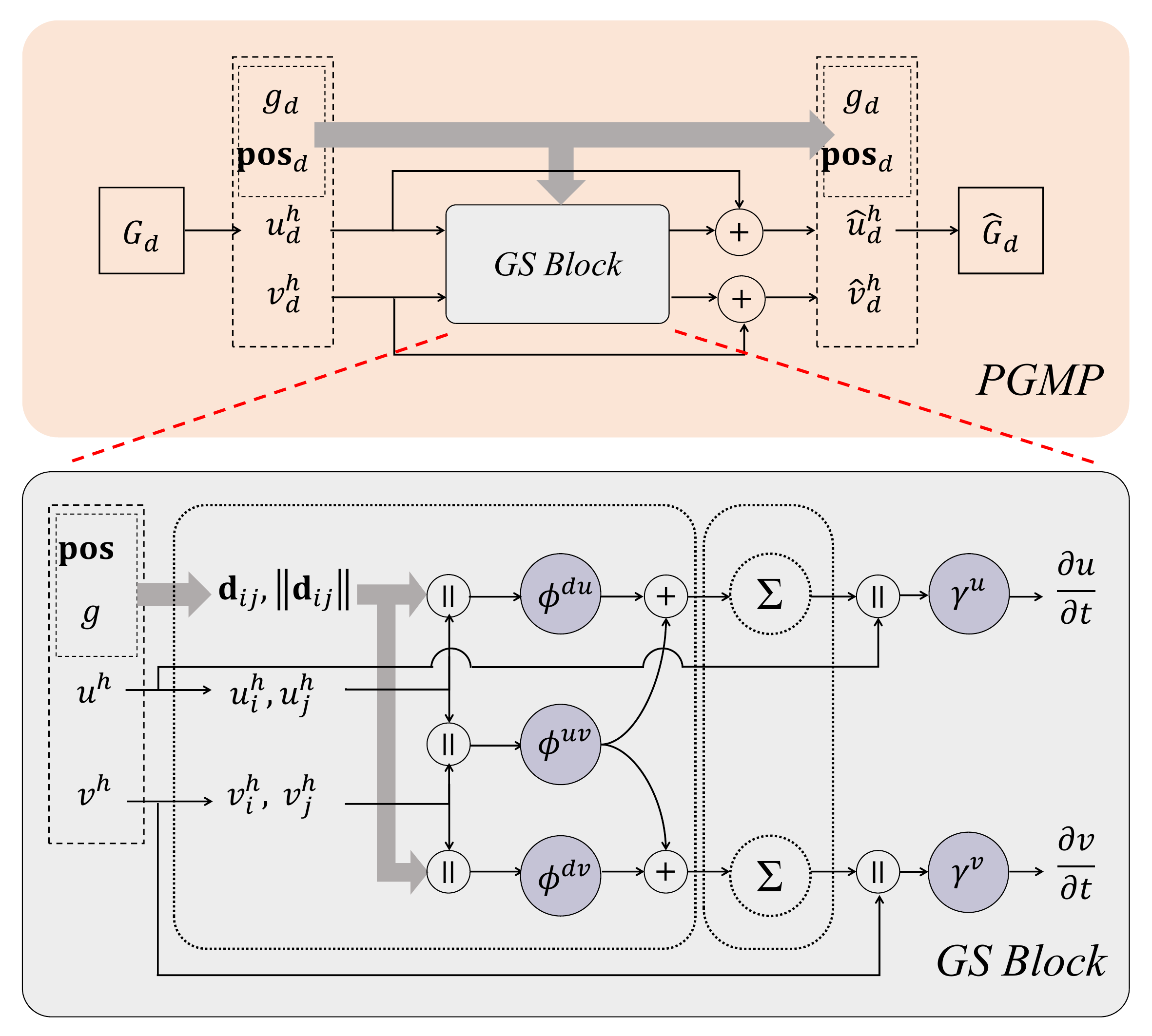}
    \caption{The specific structures of the new PGMP and the GS Block.}
	\label{fig:GSBlock}
\end{figure}

\begin{table}
	\caption{Comparison on the Gray-Scott reaction-diffusion dataset.}
	\centering
    \setlength{\tabcolsep}{1mm}

	\begin{tabular}{lccc}
	\toprule
	\textbf{Measurements} & \textbf{PEGNet(ours)} & \textbf{BSMS-GNN} & \textbf{MGN}  \\
	\midrule
	
	RMSE-1[1e-2]    &\textbf{0.89}   & 1.02  & 1.44 \\
	RMSE-50[1e-2]   &\textbf{0.39}     & 1.15   & 8.51 \\
	RMSE-100[1e-2]  &\textbf{0.59}      & 2.71    & 11.1 \\
	RMSE-1000[1e-2]  &\textbf{4.23}       & 5.67    & 15.9 \\
	RMSE-last[1e-2]   &\textbf{9.26}       & 10.2      & 35.6 \\
	\bottomrule
	\end{tabular}
	\label{tab:GS}
\end{table}

To further validate the effectiveness of our PDE-guided message passing approach on other types of PDEs, we conducted comparative experiments on the Gray-Scott reaction-diffusion dataset. Following the methodology described in Section~X and based on Equation \ref{eq:GS}, we derived a new PGMP module. The corresponding GS Block is illustrated in Figure \ref{fig:GSBlock}.
We selected BSMS-GNN and MGN as baseline models for comparison. Training details followed those used in previous experiments.
The long-term prediction results on the Gray-Scott reaction-diffusion dataset are shown in Table \ref{tab:GS}. The results demonstrate that our method remains effective when applied to other types of PDEs and effectively reduces the accumulation of errors during long-term prediction.

\section{Additional Visualizations of Experiment Results}
\label{ap:visual}

\begin{figure*}
    \centering
    \includegraphics[width=1\textwidth]{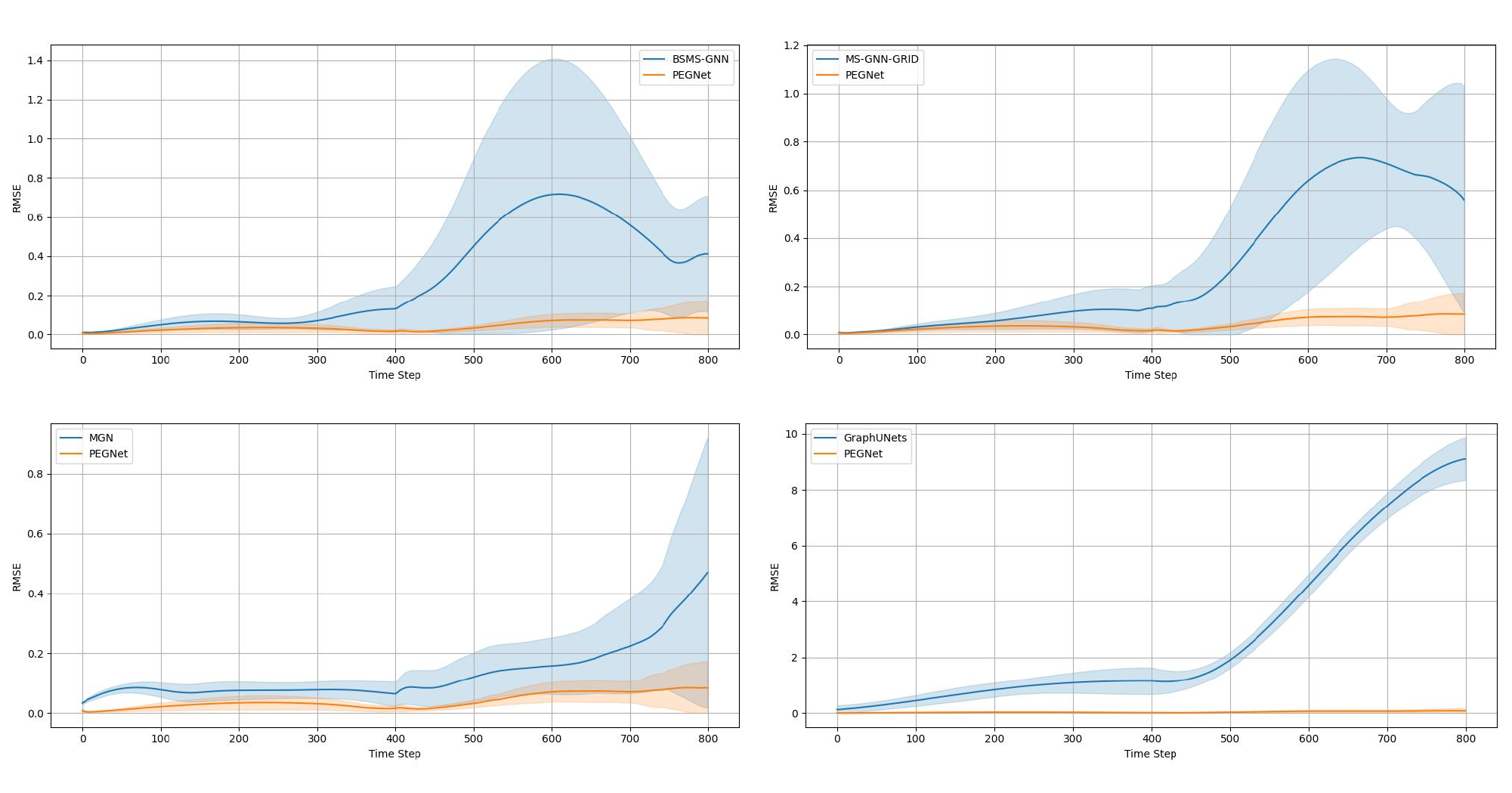}
    \caption{RMSE and its confidence interval for long-term forecasting on the Case Airflow. Across the four comparisons, our model consistently demonstrated smaller and more stable RMSE values for long-term prediction. Furthermore, the p-values from the Wilcoxon signed-rank tests were all below the 0.05 threshold, indicating statistically significant differences.}
	\label{fig:curve}
\end{figure*}

\begin{figure*}
    \centering
    \includegraphics[width=1\textwidth]{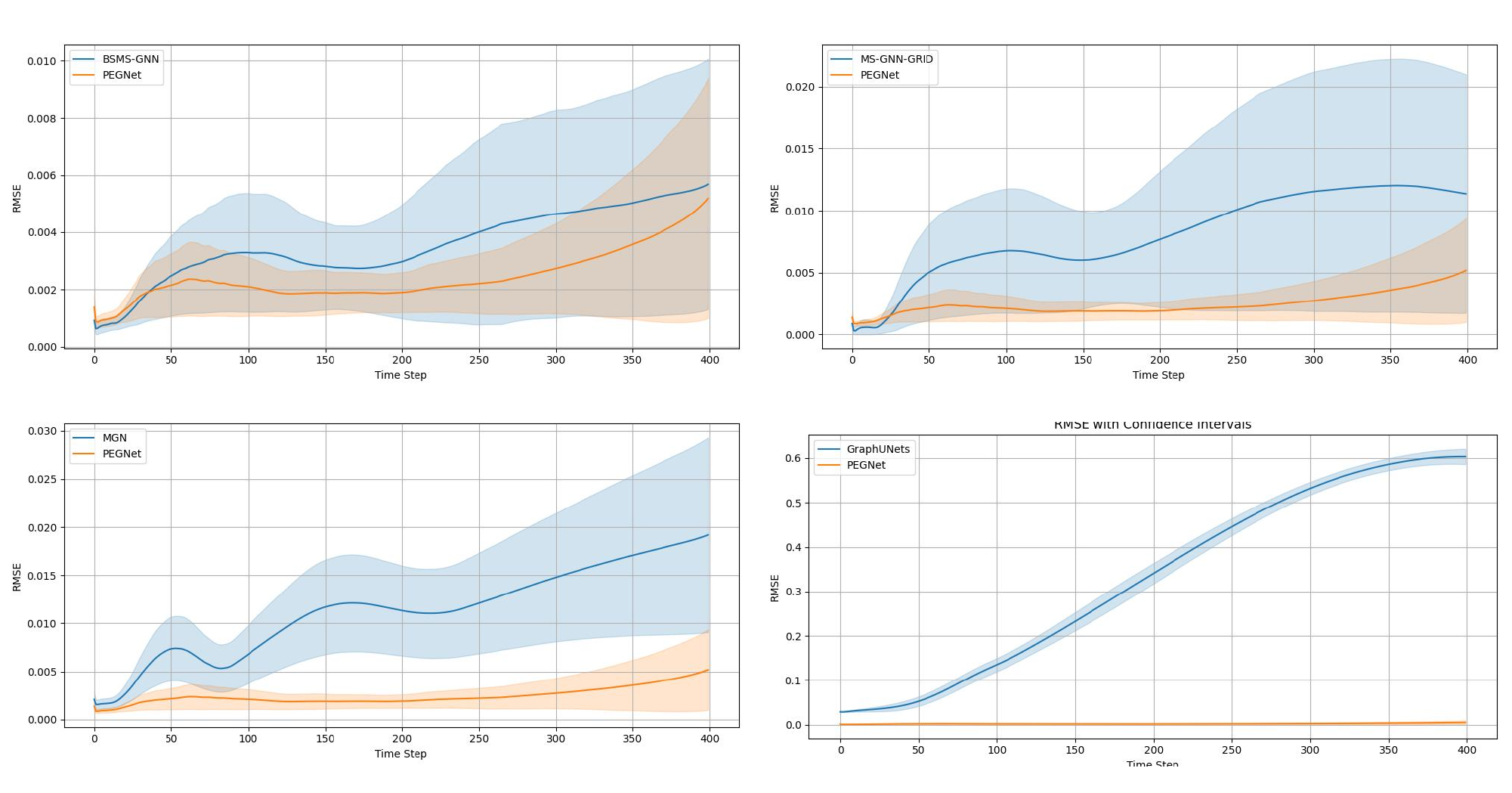}
    \caption{RMSE and its confidence interval for long-term forecasting on the Case Durg Delivery. Across the four comparisons, our model consistently demonstrated smaller and more stable RMSE values for long-term prediction. Furthermore, the p-values from the Wilcoxon signed-rank tests were all below the 0.05 threshold, indicating statistically significant differences.}
	\label{fig:curve2}
\end{figure*}

\begin{figure*}
    \centering
    \includegraphics[width=1\textwidth]{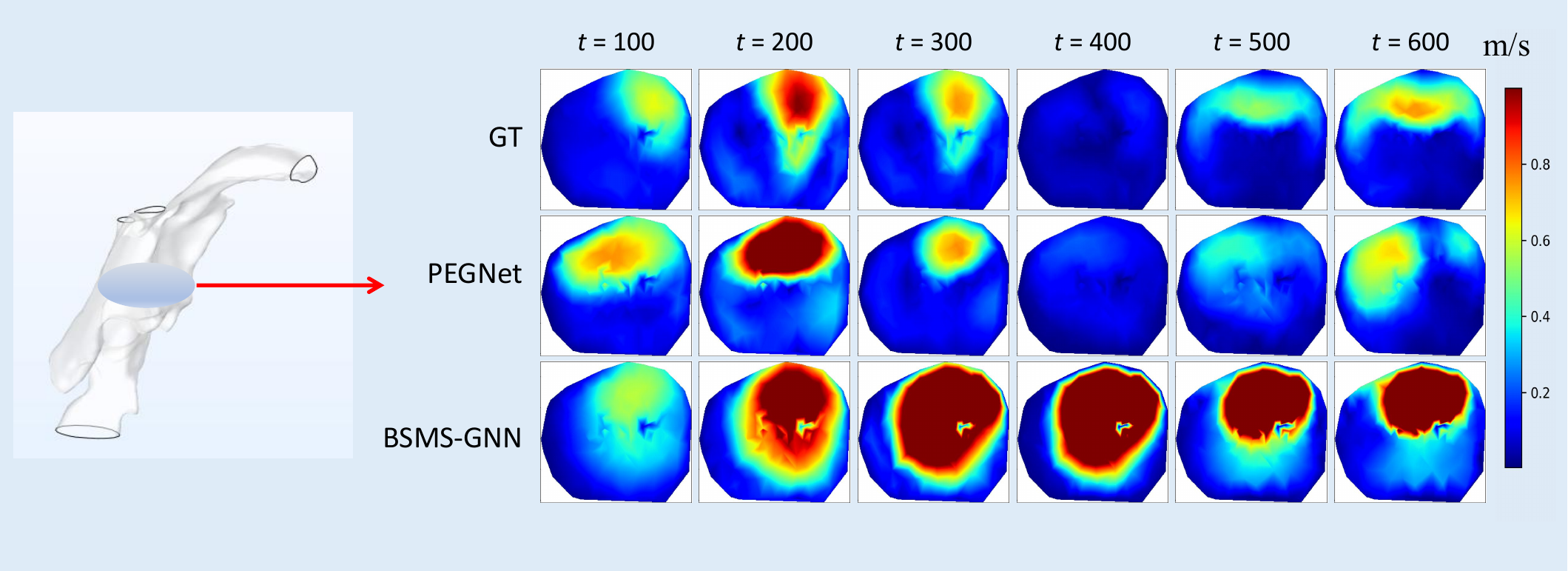}
    \caption{Comparison of airflow velocity at the midsection of the upper airway.}
	\label{fig:comparison1}
\end{figure*}

\begin{figure*}
    \centering
    \includegraphics[width=1\textwidth]{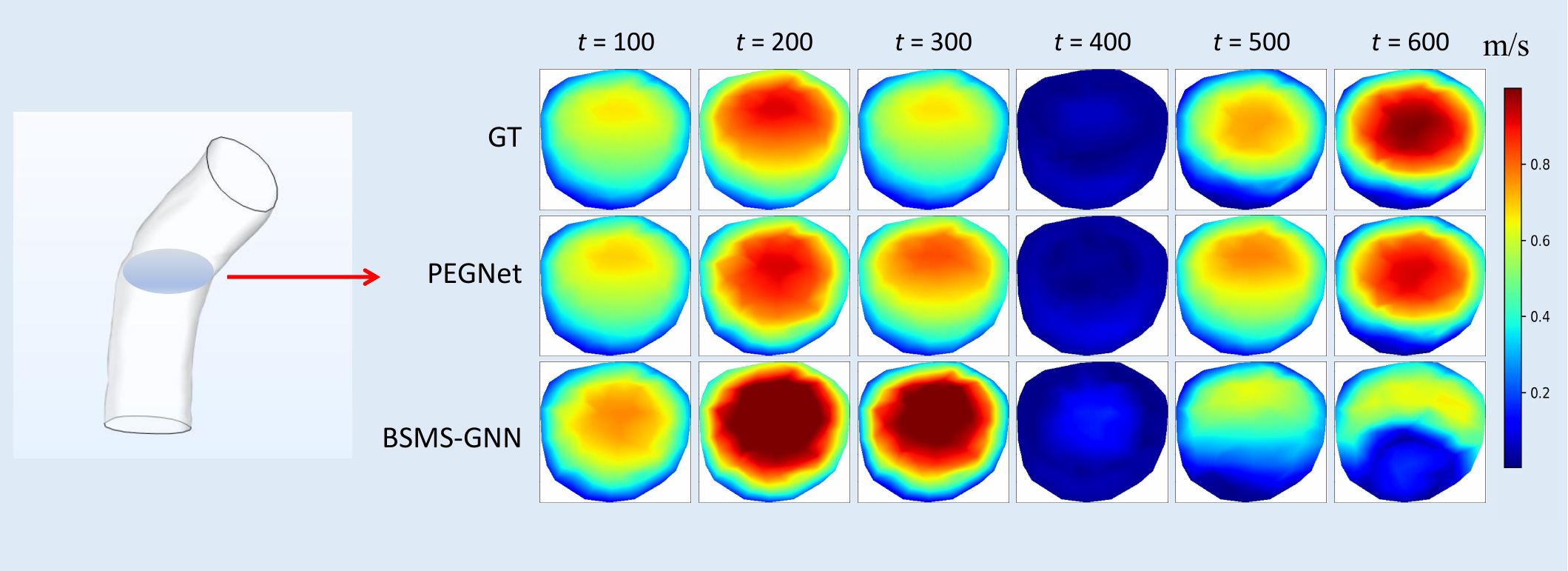}
    \caption{Comparison of airflow velocity at the midsection of the trachea.}
	\label{fig:comparison2}
\end{figure*}


\begin{figure*}
    \centering
    \includegraphics[width=1\textwidth]{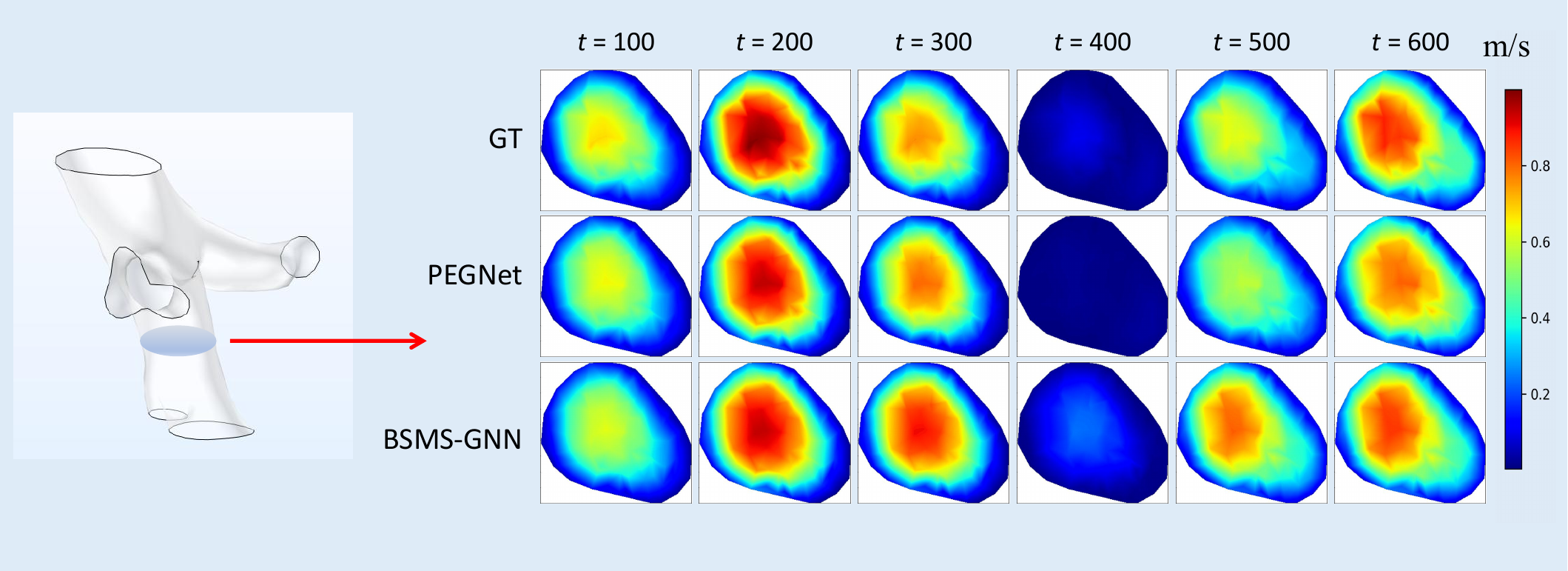}
    \caption{Comparison of airflow velocity at the end-section of the bronchi.}
	\label{fig:comparison4}
\end{figure*}

\end{document}